\documentclass[11pt]{article}

\usepackage[preprint]{acl}
\usepackage[T1]{fontenc}
\usepackage{times}

\usepackage{microtype}
\usepackage{amsmath,amssymb,mathtools}
\usepackage{booktabs,tabularx,array}
\usepackage{enumitem}
\usepackage{tikz}
\usepackage[most]{tcolorbox}
\usepackage{graphicx}
\definecolor{PaperGray}{HTML}{F4F4F4}
\definecolor{MidGray}{HTML}{6F6F6F}
\definecolor{LightGray}{HTML}{E8E8E8}
\hypersetup{
  colorlinks=true,
  linkcolor=black,
  citecolor=black,
  urlcolor=blue,
  pdftitle={Intelligence Under Time Constraints: Rethinking Test-Time Compute},
  pdfkeywords={streaming test-time compute, information-slack dilemma, metareasoning, evolving input, latency}
}

\setlist[itemize]{leftmargin=1.45em,itemsep=0.1em,topsep=0.2em}
\setlist[enumerate]{leftmargin=1.55em,itemsep=0.1em,topsep=0.2em}
\newcolumntype{Y}{>{\raggedright\arraybackslash}X}
\newcolumntype{P}[1]{>{\raggedright\arraybackslash}p{#1}}

\newtcolorbox{definitionbox}[1]{
  enhanced,breakable,
  colback=PaperGray,colframe=MidGray,
  boxrule=0.45pt,arc=0mm,
  left=1.4mm,right=1.4mm,top=1mm,bottom=1mm,
  title={#1},fonttitle=\bfseries,
  coltitle=black,colbacktitle=PaperGray
}
\newtcolorbox{patternbox}[1]{
  enhanced,
  before upper={\setlength{\parskip}{2pt}},
  colback=white,colframe=black,
  boxrule=0.55pt,arc=0mm,
  left=1.4mm,right=1.4mm,top=1mm,bottom=1mm,
  title={#1},fonttitle=\bfseries,
  coltitle=black,colbacktitle=LightGray
}
\newcommand{\term}[1]{\textbf{#1}}

\title{Intelligence Under Time Constraints:\\
Rethinking Test-Time Compute}
\author{Xiaotian Zhang \\ Trooly.AI}

\begin{document}
\maketitle

\begin{abstract}
Intelligence under time constraints requires deciding not only how much to compute, but when computation is worth starting. We study this problem in streaming interactions, where evidence arrives incrementally and may be revised. Early computation has more time to finish but rests on incomplete evidence; waiting improves information while shrinking computational slack. We call this the information--slack dilemma.

We take the evidence-dependent computational job as the unit of analysis: when to start it, what supports its result, and when that result can be committed. Advance computation is valuable only insofar as its benefits survive the costs of verification, invalidation, and recovery. This applies to grounded incremental processing and reusable preparation as well as future-dependent speculation.

We propose a research agenda on computation under evolving evidence, prioritizing selective recovery under controlled evidence revisions. Evaluation should separate earlier-execution effects, deployment value against a full-input alternative, and the added value of predictive policies, while accounting for shared-resource costs. The objective is not maximal advance computation, but more trustworthy, on-time responses within a declared resource envelope.
\end{abstract}

\section{Introduction}

A reader counting Dallas's first-quarter points can maintain a targeted tally if the question arrives before the sports passage. If it arrives last, the reader must instead prepare a general index, anticipate the question, or wait. The completed task is unchanged, but useful early work depends on its reveal schedule.

Test-time compute (TTC) commonly increases serial reasoning depth, parallel search width, or both after a complete prompt arrives \cite{muennighoff2025,wang2023,yao2023,brown2024,snell2025,wu2025,pan2025}. Interactive input instead provides a stochastic compute window. Starting early buys slack while risking obsolete work; waiting reduces uncertainty but places more computation after input completion. We call this the \term{information--slack dilemma}.

\begin{definitionbox}{Position}
Capability under streaming observation depends jointly on what an agent computes, which evidence the computation requires, and when its result is committed. Evaluation should make all three visible under a declared deadline and resource envelope.
\end{definitionbox}

We examine intelligence under time constraints through the specific setting of streaming TTC. The question is how to schedule evidence-dependent work within a response deadline. A job need not predict a continuation: grounded processing and reusable preparation also exploit input time. We develop this view into questions about waiting, state recovery, verification, and shared serving.

\subsection{An Illustrative Budgeted Example}
\label{sec:example}

Consider the following toy operator library. Four evidence blocks arrive at times $0,1,2,3$ seconds; both streams end at $T=4$. The question arrives either at $0$ or at $4$, with identical completed input. One processor has a budget of $3.5$ processor-seconds, and the complete answer is due at $4.5$ seconds. Jobs may start on input arrivals or prerequisite completion. An explicit end marker is released at $T$, with detection and transport delays ignored; policies cannot know its arrival time in advance. Answer rendering waits for this marker and the required computation.

A \emph{targeted tally} costs $0.25$ seconds per block and needs a specified question. A \emph{general index} costs $0.75$ seconds per block and supports every question in the toy family. \emph{Speculative tallying} first predicts a question in $0.25$ seconds, then runs targeted tallies; when the real question arrives, a $0.25$-second exact-match gate accepts or rejects that guess. Every strategy spends another $0.25$ seconds rendering its answer. Service is deterministic, tallies are not transferable across questions, and tallying, indexing, and the gate are exact. These are illustrative assumptions, not LLM measurements.

Table~\ref{tab:toy} summarizes the schedules. Early-question tallying finishes its last block at $3.25$; the general index finishes at $3.75$. Both wait for the end marker before rendering. Late-question tallying instead starts at $4$. Speculation with a late question must also pass its gate before rendering, or recompute all tallies on a miss.

\par\medskip\noindent
\begin{minipage}{\columnwidth}
\makeatletter\def\@captype{table}\makeatother
\centering
\small
\setlength{\tabcolsep}{3pt}
\caption{Illustrative schedules. Delivery is absolute time, not post-end latency. Work includes rendering, verification, and any recomputation; the deadline is $4.5$ s.}
\label{tab:toy}
\begin{tabularx}{\columnwidth}{Yrr}
\toprule
Policy / question arrival & Delivery (s) & Work (s) \\
\midrule
Targeted / early & 4.25 & 1.25 \\
Targeted / late & 5.25 & 1.25 \\
General index / either & 4.25 & 3.25 \\
Speculative / late, hit & 4.50 & 1.75 \\
Speculative / late, miss & 5.50 & 2.75 \\
\bottomrule
\end{tabularx}
\end{minipage}\par\medskip

At a $3.5$ processor-second budget, the early question permits cheap targeted work; with a late question, the general index is reliably on time. If a speculative guess matches with probability $0.6$, its expected on-time correctness is $0.6$, versus $1$ for the index. Reducing the budget to $2.75$ excludes the complete general-index policy, while the listed speculative policy still fits on both hits and misses and delivers on time on hits.

Thus information arrival, the deadline, and budget jointly determine useful early work. This comparison concerns the declared policy menu, not global optimality: partial indexes, hybrid policies, and other operators could change the trade-off.

\section{Intellectual Lineage}

\paragraph{Allocating computation.}
Bounded optimality relates behavior to an agent's architecture and environment \cite{russell1995}; anytime algorithms expose quality as a function of deliberation \cite{zilberstein1996}. Rational metareasoning values computation through subsequent decisions \cite{russellwefald1991}, while deliberation scheduling allocates time among procedures \cite{boddydean1994}. Continual computation addresses current and possible future problems, shared subtasks, and stale results \cite{shahaf2009}.

LLM metareasoning applies value-of-computation objectives to selective reasoning \cite{desabbata2024ram}; latency-aware TTC considers accuracy, tokens, and wall time \cite{huang2025latency}. These complete-query allocation frameworks complement search and verification operators \cite{hao2023,cobbe2021,lightman2024,setlur2025}. The question here is allocation while the specification itself evolves.

\paragraph{Revising and committing.}
Truth maintenance records justifications for belief revision \cite{doyle1979}. Incremental dialogue represents extendable and revocable information units \cite{schlangen2011}, with stability and timing measures \cite{baumann2011}. Utterance interpretation and turn prediction anticipate future input \cite{devault2011,ekstedt2021}. Simultaneous translation studies READ/WRITE decisions and latency \cite{gu2017,ma2019stacl,zheng2019adaptive,zheng2019speculative}; SimulEval supplies causal replay \cite{ma2020simuleval}. Online semantic parsing executes partial programs during utterances \cite{zhou2022}.

\paragraph{Input-time inference.}
LiveMind accumulates intermediate inferences \cite{livemind2024}; PredGen generates and verifies candidate responses during speech \cite{predgen2025}; StreamingThinker combines causal reasoning units with concurrent execution \cite{streamingthinker2026}. The streaming-LLM survey covers interaction policies and dynamic budgeting as well as architectures \cite{tong2026survey}. Our narrower focus links evidence-dependent jobs to execution, reuse, recovery, and shared-resource costs. Appendix~\ref{app:lineage} compares representative mechanisms.

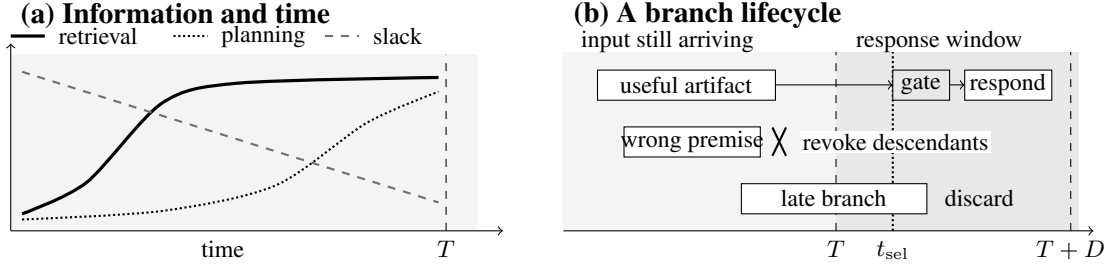
\begin{figure*}[!t]
\centering
\begin{tikzpicture}[x=1cm,y=0.75cm,font=\small]
  \node[anchor=west,font=\bfseries] at (0,4.15) {(a) Information and time};
  \fill[PaperGray] (0,0.35) rectangle (6.15,3.45);
  \draw[->] (0,0.35)--(6.5,0.35);
  \draw[->] (0,0.35)--(0,3.65);
  \draw[dashed] (5.75,0.35)--(5.75,3.45);
  \node[below] at (5.75,0.35) {$T$};
  \node[below] at (2.8,0.35) {time};
  \draw[very thick] plot[smooth] coordinates {(0.15,0.65)(1,1.2)(2,2.6)(3,2.95)(5.65,3.05)};
  \draw[thick,densely dotted] plot[smooth] coordinates {(0.15,0.55)(2,0.7)(3.5,1.15)(4.7,2.25)(5.65,2.8)};
  \draw[thick,dashed,MidGray] (0.15,3.15)--(5.65,0.85);
  \draw[very thick] (0,3.78)--(0.45,3.78);
  \node[anchor=west,font=\footnotesize] at (0.5,3.78) {retrieval};
  \draw[thick,densely dotted] (2.15,3.78)--(2.6,3.78);
  \node[anchor=west,font=\footnotesize] at (2.65,3.78) {planning};
  \draw[thick,dashed,MidGray] (4.15,3.78)--(4.6,3.78);
  \node[anchor=west,font=\footnotesize] at (4.65,3.78) {slack};
  \begin{scope}[xshift=7.3cm]
    \node[anchor=west,font=\bfseries] at (0,4.15) {(b) A branch lifecycle};
    \fill[PaperGray] (0,0.35) rectangle (3.6,3.5);
    \fill[LightGray] (3.6,0.35) rectangle (6.8,3.5);
    \draw[->] (0,0.35)--(7,0.35);
    \draw[dashed] (3.6,0.35)--(3.6,3.5);
    \draw[densely dotted,thick] (4.35,0.35)--(4.35,3.5);
    \node[below] at (4.35,0.35) {$t_{\rm sel}$};
    \draw[dashed] (6.7,0.35)--(6.7,3.5);
    \node[below] at (3.6,0.35) {$T$};
    \node[below] at (6.7,0.35) {$T+D$};
    \node[anchor=west,font=\footnotesize] at (0.1,3.68) {input still arriving};
    \node[anchor=west,font=\footnotesize] at (3.73,3.68) {response window};
    \filldraw[fill=white] (0.45,2.65) rectangle (2.8,3.18);
    \node[font=\footnotesize] at (1.62,2.92) {useful artifact};
    \draw[->] (2.8,2.92)--(4.35,2.92);
    \filldraw[fill=LightGray] (4.35,2.65) rectangle (5.1,3.18);
    \node[font=\footnotesize] at (4.72,2.92) {gate};
    \draw[->] (5.1,2.92)--(5.3,2.92);
    \filldraw[fill=white] (5.3,2.65) rectangle (6.45,3.18);
    \node[font=\footnotesize] at (5.87,2.92) {respond};
    \filldraw[fill=white] (0.8,1.65) rectangle (2.6,2.18);
    \node[font=\footnotesize] at (1.7,1.92) {wrong premise};
    \draw[thick] (2.76,1.67)--(2.94,2.17);
    \draw[thick] (2.76,2.17)--(2.94,1.67);
    \node[anchor=west,font=\footnotesize,fill=white,inner sep=1pt] at (3.12,1.92) {revoke descendants};
    \filldraw[fill=white] (2.35,0.65) rectangle (4.8,1.18);
    \node[font=\footnotesize] at (3.57,0.92) {late branch};
    \node[anchor=west,font=\footnotesize] at (4.9,0.92) {discard};
  \end{scope}
\end{tikzpicture}
\caption{Streaming TTC couples evidence sufficiency with a finite computation window. (a) The same prefix may support retrieval before it supports planning; curves are schematic, not fitted. (b) Overlapped work is not automatically useful: a premise can be revoked, a result can miss the candidate-selection cutoff $t_{\rm sel}$, and verification occupies the response critical path. This example freezes the candidate set at the cutoff; other policies may admit later results. Positions are schematic.}
\label{fig:dilemma}
\end{figure*}

\section{A Framework for Streaming TTC}

\subsection{Three Evidence-Dependency Regimes}
\label{sec:taxonomy}

We describe \emph{artifacts}, rather than models, through three non-exclusive regimes. Grounding concerns evidential support; robustness concerns usefulness across futures. A grounded artifact may also be continuation-robust, as with an index of observed events. The crucial distinction is whether applicability requires an unconfirmed future hypothesis.

\paragraph{Evidence-grounded incremental computation.}
The artifact is supported by observations already available: an entity index, a running tally, or retrieval for a stable named topic. New observations may extend it and revisions may invalidate it. Its value does not require predicting a hidden tail.

\paragraph{Continuation-robust preparation.}
The artifact is useful across a declared family of possible continuations: a reusable problem decomposition, a menu of neutral follow-up dimensions, or a general event index. It can be unused or insufficient without being factually wrong. Robustness must be tested over that family; a generic checklist is not useful merely because it is safe.

\paragraph{Future-dependent speculation.}
The artifact's applicability depends on an unresolved hypothesis: a guessed intent, missing constraint, or likely answer. It must remain conditional until supported or rejected. Prediction coverage and false acceptance matter specifically here.

These regimes cut across operator types: retrieval or planning can belong to any of them. A note can contain fields from several regimes, which should be tracked separately. Predictable future content is one route to positive input-time value, not a necessary condition for the entire framework.

\subsection{Episodes, Observations, and Clocks}

Let $T$ be an almost-surely finite \emph{source-end event}: actual user stop-speaking time, or the last source release in a text replay. Let $X$ be the completed, canonically serialized task input. The agent receives observations $O_{\le t}$, including partial transcripts, revisions, and stability signals. Their history generates $\mathcal I_t$; the controller filtration $\mathcal F_t\supseteq\mathcal I_t$ also includes jobs, artifacts, queues, and resource events.

In speech, distinguish $T$ from the detected endpoint $T_{\rm det}$ and finalized-transcript availability $T_{\rm fin}$. The evaluator may know $T$ retrospectively; it is not generally a stopping time of the agent's observation filtration. The controller uses a causal posterior, not the annotated endpoint. The core protocol allows response commitment only after the episode ends, with premature commitment scored separately as a violation. Continual action before source completion requires a different deadline process.

\paragraph{Evidence sufficiency is operator-specific.}
A stable entity may suffice for retrieval before the user's intent suffices for planning. Scheduling nevertheless requires three separate assessments: whether the current evidence supports a job, its completion-time distribution, and the downstream value of its result. No single maturity score is assumed sufficient. Appendix~\ref{app:maturity} gives population Bayes risk as one optional characterization, not a required controller statistic.

For delay allowance $D$, realized remaining deadline slack is $(T+D-t)_+$. A controller estimates remaining time from causal observations; for example, $\widehat S_t=\mathbb E[(T-t)_+\mid\mathcal F_t]$ estimates expected pre-end slack. This mean alone does not characterize deadline-miss risk, which depends on the joint distribution of remaining input time, job service, and queues. Endpoint detection, transcript finalization, and verification may consume part of $D$.

\subsection{Causal Jobs and Resource Constraints}

A state $h_t$ summarizes observed evidence, operator-specific evidence estimates, the endpoint posterior, running jobs, queues, artifacts, and capacity. The action space is
\begin{equation}
a_t\in
\left\{
\begin{array}{l}
\operatorname{wait},\ \operatorname{spawn}(o,c,P),\\
\operatorname{continue}(j,c'),\ \operatorname{cancel}(j),\\
\operatorname{verify}(j,c_v),\ \operatorname{commit}(y)
\end{array}
\right\},
\label{eq:actions}
\end{equation}
where $c$ is a budget and $P$ contains prerequisites. Operators include encoding/prefill, retrieval, extraction, planning, branching, verification, and response generation.

Each job records its operator, launch time, evidence version, dependency set, service demand, and output artifact. A frozen-snapshot job uses only its recorded launch prefix. An online-update job may consume later observations, but every update must be measurable with respect to $\mathcal F_t$ at its execution time. Both are causal; a retrospective final transcript is not an admissible launch input.

If job $j$ receives service $\alpha_j(t)$ with demand vector $\mathbf r_j$, capacity requires
\begin{equation}
\sum_{j\in\mathcal J_t}\alpha_j(t)\mathbf r_j
\preceq\mathbf K(t).
\label{eq:capacity}
\end{equation}
Cancellation must include the time until a backend actually stops charging resources. Deleting a client-side handle does not imply reclaimed capacity.

\subsection{Quality, Deadlines, and Critical Paths}

Fix a task-specific response endpoint $A_\pi$ and score the content $Y_\pi$ available by that endpoint. It may be the complete answer-bearing response or a substantive interview follow-up. Set $L_\pi=A_\pi-T$ for a post-end response, and $L_\pi=\infty$ if none is produced. Later content must not improve the score retroactively. For quality $q\in[0,1]$, we take on-time quality as the primary objective:
\begin{equation}
J_D(\pi)=\mathbb E\!\left[
q(Y_\pi;X,\Theta)\mathbf1[L_\pi\le D]\right].
\label{eq:objective}
\end{equation}
We maximize $J_D$ over causal policies subject to resource and risk limits. Resource limits bound expected consumption $\mathbb E[C_{\pi,k}]\le B_k$ and instantaneous capacity in Equation~\eqref{eq:capacity}; an application may additionally bound external harm. Gates, cancelled work, and state reconstruction count toward consumption. A launch is valuable when its expected gain in deadline utility exceeds that of waiting or another feasible use of the same resources, including effects on later decisions.

Maximizing eventual quality subject to a deadline-miss tolerance is a different objective. With nonzero allowed misses it can rank policies differently from $J_D$, so the two must not be used interchangeably. Raw quality, latency, and cost remain necessary to interpret either.

Work overlapped with input is not latency saved. Response delay depends on the remaining dependency path, including detection, finalization, verification, and serving. Unused branches can still impose contention until stopped. Overlapped branch-seconds cannot be summed into response acceleration; Appendix~\ref{app:critical} gives a residual-work lower bound.

\section{Research Agenda: Computation under Revisable Evidence}
\label{sec:agenda}

The agenda is to predict when a job has positive value and to make its intermediate state safe to reuse. Time gained by starting early can be consumed by later validity checks and state reconstruction. Recovery cost therefore belongs in the launch decision itself: it is a consequence of the information--slack dilemma, not merely a maintenance concern. Each question below pairs a design problem with a discriminating experiment.

\subsection{When Is Waiting Worth More than Computing?}

Input progress conflates evidence sufficiency with remaining time. A late prefix may identify the task but leave too little time for a useful job. Inserting a pause into an otherwise identical stream creates additional computation time before source end; within a realized episode, slack still decreases as time passes. A pause can also change the endpoint posterior.

Vary informative-segment arrival independently of pauses, keeping completed input and resources fixed. Compare triggers based on online-estimated progress, evidence alone, slack alone, and their combination. True-final-length percentages are oracle diagnostics. A joint controller should improve deadline utility in mixed regimes; if evidence and time estimates add no predictive value beyond elapsed time, the richer state is unnecessary.

\subsection{What Work Remains Useful across Futures?}

A general event index can support several eventual questions without predicting any. A correctly guessed topic can nevertheless produce a useless plan. An artifact's \emph{reuse region} consists of terminal episodes in which retaining it improves the downstream decision under a fixed response budget.

Construct groups sharing early evidence but differing in withheld questions, constraints, or intentions. Compare the three regimes in Section~\ref{sec:taxonomy} on utility and invalidation, not only textual match. Future-dependent preparation earns a distinct claim only by improving on strong non-predictive processing. Unpredictable tails do not eliminate reusable shared subproblems.

\subsection{When Is Local Retraction Enough?}

Deleting an assumption from a note does not necessarily remove its influence. Two dependency regimes call for different recovery strategies.

\paragraph{Explicit dependence.}
A derived claim records the assumption or evidence version supporting it. Following truth maintenance and incremental-unit models \cite{doyle1979,schlangen2011}, local invalidation can remove unsupported descendants, cancel dependent jobs, and retain claims with independent surviving justifications. This is sound only to the extent that the recorded support captures actual dependence.

\paragraph{Implicit influence.}
A summary, later generation, or model cache may have absorbed a premise without retaining its provenance. Removing its source may leave unsupported influence in downstream state. Recovery may require rebuilding from independently supported evidence or a checkpoint preceding exposure to that premise. Reconstruction rebuilds computational state from valid evidence; it does not guarantee a correct final answer. A text label does not certify that a cache is clean.

The research question is therefore when local invalidation is sufficient and when reconstruction is worth its cost. A controller might use artifact isolation, provenance coverage, state compression events, and contradiction severity as risk signals, rather than assume a complete dependency graph. Evaluate selective recovery against always-local and always-rebuild policies using controlled corrections. Measure stale-claim survival, valid-work retention, recovery latency, and total cost. The target is reliable selection of recovery scope, not merely a more elaborate note format.

\subsection{When Does Verification Pay for Itself?}

Compatibility, utility, and readiness differ. A supported branch may add nothing the final model needs; a useful branch may arrive too late. Gate runtime, added context, and regeneration can consume the gain.

Use paired accept/drop interventions on a candidate and final input. Compare similarity-only, support-aware, and utility-aware gates, alongside always-inject and never-inject. Measure correct-to-wrong flips and critical-path change, including gate cost. Offline utility labels can train selection, but future evidence remains unavailable to the online gate.

\subsection{Whose Latency Is Being Reduced?}

Early jobs can improve one user's response while delaying another. Compare admission, cancellation, and foreground-priority policies under fixed workload and hardware traces. Measure neighboring requests, throughput, tail latency, and resource use as load increases.

A deployment win should persist within a declared load envelope. The controller should throttle early work when opportunity cost exceeds benefit. Waiting is then a rational allocation decision, not failed prediction.

\paragraph{Where to start.}
We prioritize controlled replays with evidence revisions, comparing selective reconstruction against always-local invalidation and always-full reconstruction under matched resources. Retain a no-semantic-precomputation control that solves from the complete input, as in Section~\ref{sec:eval}. Unlike always-full reconstruction, this control incurs neither advance semantic work nor its queueing effects. It separates better repair from the value of adopting advance computation and repair at all. Vary revision timing, dependency visibility, and the extent to which summaries or caches absorb revoked premises. Measure on-time quality, stale-claim survival, and recovery cost. These experiments can establish when retained state remains usable and what repair costs before a learned scheduler is asked to allocate work around it.

\section{Evaluating Evidence-Dependent Computation}
\label{sec:eval}

\subsection{Match the Control to the Claim}

Let $\pi$ be any streaming policy under evaluation. Three contrasts answer different questions (Table~\ref{tab:claims}); none requires that $\pi$ predict future input or launch multiple waves.

\begin{table*}[t]
\centering
\small
\renewcommand{\arraystretch}{1.12}
\caption{Three claims about an arbitrary streaming policy $\pi$. Timing, deployment, and prediction benefits require distinct comparisons.}
\label{tab:claims}
\begin{tabularx}{\textwidth}{P{3.0cm}P{5.0cm}Y}
\toprule
Claim & Required comparison & Scope of the conclusion \\
\midrule
Earlier execution helps & Freeze a computational trace and intervene on job release times & Scheduling value conditional on the recorded jobs and evidence \\
A streaming policy has deployment value & Compare with a specified full-input policy under the same resource and load envelope & Improvement over that alternative, not every possible post-input algorithm \\
Predicting future input adds value & Compare with strong incremental processing and continuation-robust preparation & Benefit of the predictive policy package; a pure assumption effect needs tighter ablation \\
\bottomrule
\end{tabularx}
\end{table*}

A \emph{trace-conditioned timing control} preserves the recorded job graph, evidence snapshots, and service demands in simulation, while delaying early releases. It tests whether overlap shortens the response path. Recomputing jobs from full evidence changes the information condition and answers a different question. Appendix~\ref{app:replay} specifies one intervention and distinguishes frozen-trace simulation from stochastic live re-execution.

A \emph{full-input control} waits for complete evidence and uses a declared allocation rule within the same budget and serving envelope. Restricting its operator portfolio aids controlled comparison but narrows the deployment claim. A practical study should also compare the strongest relevant deployed alternative; a win over one chosen portfolio does not establish global superiority.

A \emph{non-predictive control} performs evidence-grounded processing and robust preparation. If its prompts, selection rules, or artifact types differ, the comparison estimates the benefit of the whole predictive policy, not isolated future assumptions. Prefill, native thinking, and single-wave controls are useful for specific additional claims and are listed in Appendix~\ref{app:controls}.

\subsection{Make Information and Time Observable}

Pair policies on completed inputs, timestamped arrivals, model versions, output caps, and serving conditions. Include controlled reveal schedules that change question availability while preserving final serialization, as in Section~\ref{sec:example}, and pause interventions that preserve evidence order. Naturally occurring streams test whether the findings transfer. Future input length and annotated source end are evaluator-only information.

Measure response latency from actual source end $T$, while logging detection $T_{\rm det}$ and transcript finalization $T_{\rm fin}$ separately. First token, first audio, first substantive audio, availability of the scored response, and task completion are distinct endpoints. Report their distributions as relevant, rather than treating them as interchangeable.

Estimate $J_D$ by averaging quality multiplied by the indicator that the scored response was available by its deadline. Alternatively, evaluate the actual content delivered by $T+D$ under an incremental rubric. Neither convention credits \emph{Let me think} at 100 ms with the quality of an answer delivered ten seconds later. Raw quality, timeout/error counts, and total compute and monetary cost should accompany deadline utility.

\subsection{Interpret Gains at the Right Level}

Preregister the main endpoint, deadline, resource envelope, and quality non-inferiority margin. Use paired comparisons with case-level uncertainty estimates; seeds and reveal schedules from one source case are not independent samples. Mechanism metrics should include supported-artifact reuse, stale-claim survival, correct-to-wrong flips, recovery cost, and the residual critical path. Interview harm rubrics should cover leading questions based on unsupported premises.

A streaming policy can succeed by preserving the quality of a stronger reasoning system while approaching the latency of a direct response. It need not exceed native-thinking accuracy. Conversely, an apparent speedup is insufficient if it trades away unreported quality, spends substantially more resources, or delays neighboring requests. Benefits should be attributed at the level identified by the relevant control: incremental processing, future-dependent preparation, or a particular scheduling mechanism.

\section{Trooly as an Illustrative Design Pattern}

Trooly's interviewer motivates private preparation during speech, aiming for a useful follow-up within a short response window. After \emph{I mainly use Kimi for customer reports}, a processor can record the use case, prepare neutral workflow questions, or speculate that raw-data organization is the bottleneck.

Confirmation may make a data-related plan reusable. But after \emph{The data is already organized; I only use it to adjust the tone}, its premise should be revoked while the confirmed use case survives. Asking \emph{How much cleaning time did it save?} then illustrates false acceptance: topical similarity conceals an unsupported assumption. If only an isolated branch depends on that premise, local invalidation can suffice. If a summary or cache absorbed it, the response context may need reconstruction.

The design pattern is to launch from causal evidence, track support and assumptions, invalidate obsolete work, and select useful artifacts before committing a response. Selection deadlines and protected response resources determine whether even a valid artifact can still help.

\paragraph{Evidence status.}
The scheduling example and interview alternatives are analytical illustrations. This paper proposes a problem and research agenda, not a measured performance improvement. Typed notes motivate the design; they do not establish complete dependency tracking or reliable recovery.

\section{Objections and Scope}

\paragraph{Will faster models make the problem disappear?}
They can eliminate its practical importance for some task--deadline pairs. If a full-input policy already meets the target quality, cost, and latency, anticipatory computation may add no value. The agenda concerns regimes where useful computation remains on a constrained response path; it predicts that some regimes will vanish as inference improves.

\paragraph{Why metareasoning if simple incremental processing works?}
A fixed incremental procedure may be the best policy when evidence is stable and its reuse is predictable. An adaptive controller earns its complexity only when selecting jobs, waiting, or changing recovery scope improves outcomes after controller overhead is counted. Metareasoning defines the allocation question; it does not require a learned scheduler in every deployment.

\paragraph{What is added beyond existing streaming research?}
Continual computation already studies advance work and shared subproblems; the streaming-LLM survey explicitly covers interaction policies and runtime budgeting \cite{shahaf2009,tong2026survey}. Our proposed contribution is a narrower job-level research object linking evidence dependence to execution, reuse, and revocation, with controls separating timing from information and prediction effects. The test of this framing is whether it yields tractable questions such as selective state reconstruction, not whether it creates a new label for streaming architectures.

We restrict the model to episodic streams with an uncertain source end and post-end response deadline. Continuous action before source completion needs a sequence of decision opportunities. Inferring semantic dependence and calibrating recovery risk remain open, rather than assumed capabilities.

\section{Conclusion}

For streaming agents, making better use of test-time compute means choosing which work to start, reuse, verify, or rebuild before a deadline. Incremental processing, reusable preparation, speculation, and waiting are alternative ways to manage evolving evidence. An evidence-dependent job can finish early yet become invalid, or remain valid yet arrive too late to help. Its value therefore depends on both its contribution to the response path and the cost of trusting or rebuilding its result. The objective is trustworthy, on-time responses within a resource budget, rather than maximal advance computation. The question is which computations are worth doing now, and which results remain worth relying on as evidence changes.

\appendix
\section{Representative Mechanisms}
\label{app:lineage}

\begin{table*}[t]
\centering
\small
\renewcommand{\arraystretch}{1.1}
\caption{Representative contributions inherited by the agenda. Entries describe each mechanism's focus; they do not assert that unlisted capabilities are absent.}
\label{tab:lineage}
\begin{tabularx}{\textwidth}{P{2.35cm}P{3.15cm}P{3.35cm}Y}
\toprule
Work & Computation object & Evidence and commitment & Control / objective \\
\midrule
Russell--Wefald (1991) & Internal computations affecting an external decision & Evaluate a computational result before choosing action & Select computation by expected net decision value \\
Shahaf--Horvitz (2009) & Current/future problems and shared subtasks & Precompute reusable results; consider result freshness & Allocate time under uncertain arrivals and utility \\
Doyle (1979) & Beliefs with recorded justifications & Revise assumptions and their supported beliefs & Dependency-directed belief maintenance \\
STACL (2019) & Translation prefix & Read partial source, then emit target tokens & Control quality--latency trade-off through prefix lag \\
LiveMind (2024) & Intermediate inferences on incoming segments & Store and reuse inference memory as input grows & Decide whether to infer or await more input \\
PredGen (2025) & Candidate response and speech preparation & Verify/revise candidates before output & Hide response preparation within speech input \\
StreamingThinker (2026) & Streaming reasoning units and KV execution & Causal reading/reasoning; final answer follows & Concurrent execution and post-reading depth adjustment \\
Huang et al. (2025) & Reasoning method for a complete query & Allocate inference after query receipt & Select method using accuracy, tokens, and wall time \\
\bottomrule
\end{tabularx}
\end{table*}

Table~\ref{tab:lineage} compares specific mechanisms along evidence, commitment, and control dimensions. It is a selective intellectual map, not a systematic literature review or a completeness ranking.

\section{One Candidate Measure of Evidence Sufficiency}
\label{app:maturity}

For a fixed operator $o$, let $Z_o^\star=\psi_o(X,\Theta)$ be a terminally relevant artifact and $\ell_o$ its mismatch loss. One population measure of evidence sufficiency uses Bayes risk:
\begin{equation}
\mathcal R_o(t)=
\mathbb E\!\left[
\inf_{z\in\mathcal Z_o}
\mathbb E[\ell_o(z,Z_o^\star)\mid\mathcal I_t]
\right].
\end{equation}
Let $\mathcal R_o^\infty$ be the risk after all episode observations and revisions arrive, excluding later episodes. If $\mathcal R_o(0)>\mathcal R_o^\infty$, normalize:
\begin{equation}
M_o(t)=
\frac{\mathcal R_o(0)-\mathcal R_o(t)}
     {\mathcal R_o(0)-\mathcal R_o^\infty}.
\end{equation}
Nested observation histories imply non-increasing population risk for a fixed loss, even when transcript text is corrected. Individual estimates can fluctuate. This information measure encodes neither runtime nor downstream utility.

Robust preparation may admit many equally useful representations. A loss allowing equivalent artifacts, or a downstream-regret measure, may be more appropriate. These equations are a candidate characterization, not the definition of the dilemma.

\section{Additional Mechanism-Specific Controls}
\label{app:controls}

Choose additional controls according to the mechanism claimed:
\begin{itemize}
\item \textbf{Direct response:} an ordinary no-semantic-precomputation speed/quality anchor.
\item \textbf{Native reasoning:} a serial-reasoning anchor, with measured cost and disclosed limits on hidden-budget control.
\item \textbf{Streaming prefill:} causal encoding and genuine KV reuse, without semantic artifacts. Repeated full-prefix API calls are not equivalent.
\item \textbf{Single wave:} for multi-wave claims, a causally selected single launch under the same budget envelope. True-final-length triggers are oracle diagnostics.
\item \textbf{Recovery and gates:} always-local versus always-rebuild recovery; always-inject and never-inject versus selective reuse.
\end{itemize}

Match resource caps and final generation settings; report actual consumption, including discarded jobs. These are implementation-specific controls.

\section{Replay Manifest and Trace Requirements}
\label{app:replay}

For frozen-snapshot jobs with release times $r_j$, a timing intervention can set
\begin{equation}
r'_j=\max(T,r_j).
\label{eq:release}
\end{equation}
Prerequisites and queues still govern execution. This clamping intervention differs from a uniform time shift and must be preregistered.

Preserve job inputs, outputs, consumed service demands, priorities, and cancelled work. Keep exogenous neighboring arrivals fixed and recompute endogenous queues. Freezing outputs makes this a trace-conditioned scheduling diagnostic, not a new policy evaluation. Online-update jobs require recorded evidence-consumption steps. Live APIs that cannot reproduce service or cancellation behavior require replicated-run uncertainty rather than claims of exact isolation.

An auditable trace links timestamped evidence versions and revisions to job prompts, dependencies, budgets, and queue/start/end/cancellation events. It records artifact support, invalidation, selection, resource use, and all scored response events: $T$, $T_{\rm det}$, $T_{\rm fin}$, audio onset, substantive response, and completion. Future schedules and annotations remain evaluator-only information.

\section{Residual Response-Time Accounting}
\label{app:critical}

Let $G_\pi$ be the realized dependency DAG needed for the scored response, including detection, finalization, gates, and serving dependencies. With residual critical path $\operatorname{CP}_T(G_\pi)$ and required residual work $W_{k,T}$ on resource $k$, constant post-end capacity implies
\begin{equation}
L_\pi\ge
\max\left\{
\operatorname{CP}_T(G_\pi),\
\max_k W_{k,T}/K_k
\right\}.
\label{eq:makespan}
\end{equation}
Unused branches are not response dependencies, but still impose contention until stopped. Queueing and scheduling can further increase latency. Overlapped branch-seconds are an accounting measure; they cannot be summed into response acceleration.

\end{document}